\documentclass[letterpaper, 10 pt, conference]{ieeeconf}  

\IEEEoverridecommandlockouts                              

\usepackage{graphics} 
\usepackage{epsfig} 
\usepackage{mathptmx} 
\usepackage{times} 
\usepackage{amsmath} 
\usepackage{cite}
\usepackage{amssymb}  
\usepackage[T1]{fontenc}
\usepackage{stix}
\usepackage{booktabs}
\usepackage{multirow}
\usepackage{graphicx}
\usepackage{amsfonts,amssymb} 
\usepackage{float}

\makeatletter
\let\IEEEcaption\@makecaption
\makeatother
\usepackage[font=footnotesize]{subcaption}
\makeatletter
\let\@makecaption\IEEEcaption
\makeatother

\let\labelindent\relax
\usepackage{enumitem}

\usepackage{color}

\newtheorem{definition}{Definition}

\newtheorem{remark}{Remark}[definition]

\usepackage{macros/logic}

\title{\LARGE \bf
Safe and Personalizable Logical Guidance for Trajectory Planning of Autonomous Driving
}

\author{Yuejiao Xu$^*$, Ruolin Wang$^*$, Chengpeng Xu, and Jianmin Ji$^\dagger$
    \thanks{Yuejiao Xu, Ruolin Wang, Chengpeng Xu, and Jianmin Ji are with the University of Science and Technology of China, Hefei 230026, China.}
    \thanks{$*$ These authors contributed equally.}
    \thanks{$\dagger$ Jianmin Ji is the corresponding author {\tt\small jianmin@ustc.edu.cn}.}%
}

\begin{document}

\maketitle
\thispagestyle{empty}
\pagestyle{empty}


\begin{abstract}

Autonomous vehicles necessitate a delicate balance between safety, efficiency, and user preferences in trajectory planning.
Existing traditional or learning-based methods face challenges in adequately addressing all these aspects. 
In response, this paper proposes a novel component termed the Logical Guidance Layer (LGL), designed for seamless integration into autonomous driving trajectory planning frameworks, specifically tailored for highway scenarios.
The LGL guides the trajectory planning with a local target area determined through scenario reasoning, scenario evaluation, and guidance area calculation.
Integrating the Responsibility-Sensitive Safety (RSS) model, the LGL ensures formal safety guarantees while accommodating various user preferences defined by logical formulae. 
Experimental validation demonstrates the effectiveness of the LGL in achieving a balance between safety and efficiency, and meeting user preferences in autonomous highway driving scenarios.

\end{abstract}


\section{Introduction}
Ensuring safety is paramount for autonomous vehicles,
and it is also inevitable to consider user preferences~\cite{park2020driver}.
However, due to the diverse driving scenarios, balancing safety, efficiency, and user preferences is challenging with existing trajectory planning approaches.
Traditional sampling- or optimization-based approaches measure the comfort and safety of a trajectory through a cost function, but user preferences for driving styles go far beyond this.
Learning-based approaches can learn from human driving data, but the explicit handling of safety constraints remains an open problem.

Formal methods have been widely used for the safety guarantee of autonomous vehicles due to their high degree of interpretability and verifiability~\cite{shalev2017formal,hilscher2013proving,schwammberger2018abstract,wang2023a2cost}.
However, existing methods are mainly used as additional safety checks.
Several research attempts to integrate formal methods, especially logical-based methods, into the planning framework~\cite{wongpiromsarn2012receding,qian2016optimal,hubmann2017decision,sahin2020autonomous}.
However, they are tied to a specific planning method or fall short of bridging the gap from logical discrete to continuous space.

\begin{figure}[t]
    \centering
    \includegraphics[width=0.9\linewidth]{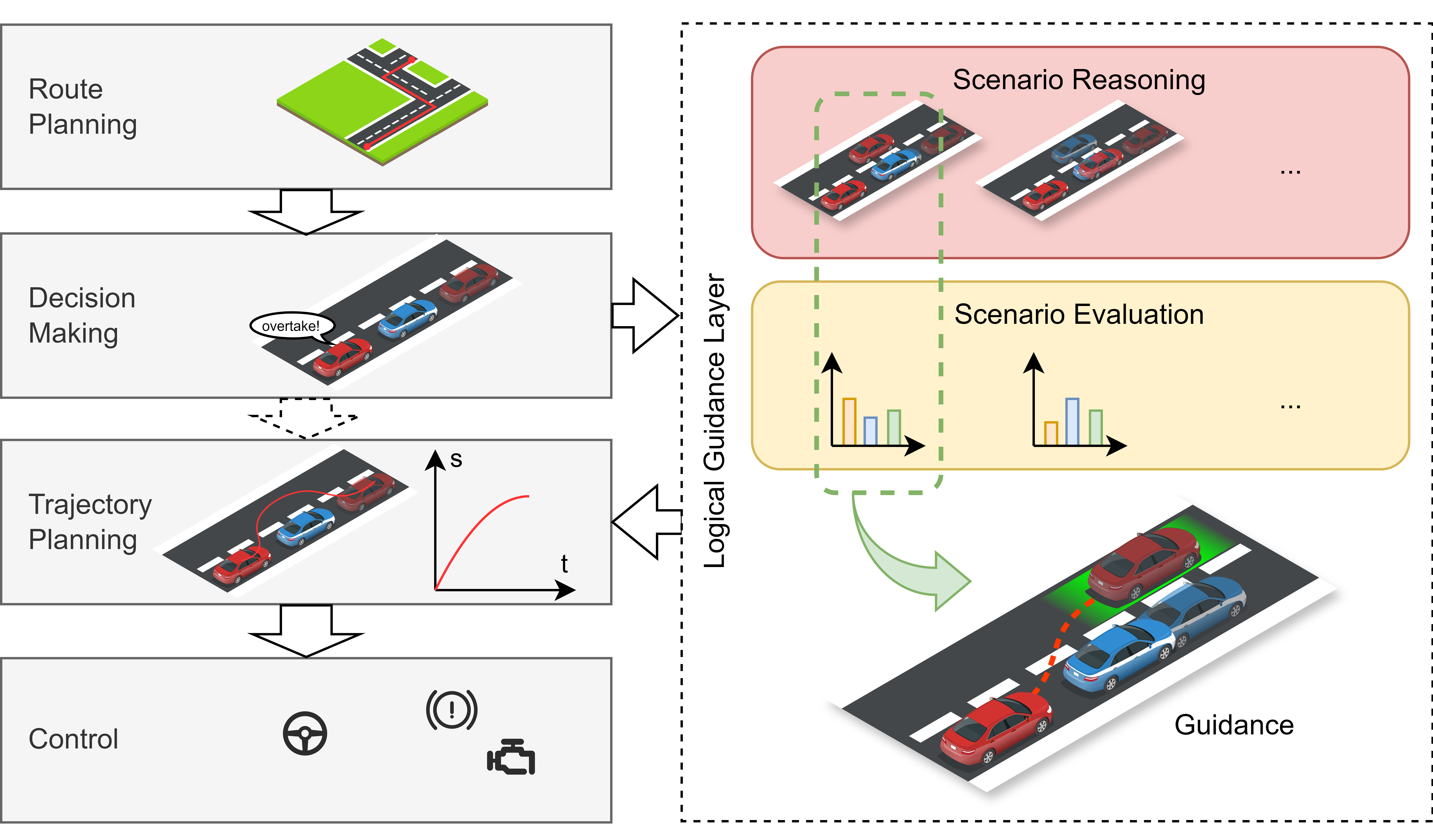}
    \caption{The role of the proposed Logical Guidance Layer in a typical planning and control framework.
    The LGL performs as a middle layer between decision-making and local trajectory planning.
    }
    \label{fig:keynote}
\end{figure}

In this paper, we propose a novel component called \textbf{Logical Guidance Layer (LGL)},
which can be straightforwardly integrated into most of the trajectory planning frameworks of autonomous driving for highway scenarios,
with consideration of safety, efficiency, and user preferences.
The role of LGL is shown in Fig.~\ref{fig:keynote}.
Instead of initiating the trajectory planning directly based on the goal given by the decision-making layer,
the LGL guides it with a local target area.
The LGL uses scenario reasoning to discover \textit{logical scenarios} that satisfy the safety constraints and can achieve the goal.
One of these logical scenarios is selected by its efficiency and user's preferences,
and projected to the continuous space to generate the local target area.

In summary, the contributions of this paper are:
\begin{itemize}
\item We propose a novel Logical Guidance Layer that can be easily integrated into the trajectory planning framework with extra personalization ability.
\item We offer the Highway Traffic Scenario Logic (HTSL) for the scenario reasoning, with RSS model~\cite{shalev2017formal} integrated to provide a formal and verifiable safety guarantee.
\item We introduce a user preference priority setting based on logical formulas in our proposed method, which allows a high degree of freedom for personalization.
\end{itemize}

\section{Related Work}

\subsection{Logic of Traffic Scenarios}

Some existing work has attempted to build logical representations of traffic scenarios.
Esterle et al.~\cite{esterle2019} formalized traffic rules under Linear Temporal Logic semantics and developed a runtime maneuver verification strategy.
Maierhofer et al.~\cite{maierhofer2022formalization} further extended the logical representation of traffic rules.
However, such methods make it difficult to represent dynamic traffic scenarios.

Alternatively, Hilscher et al.~\cite{hilscher2011abstract} introduced Multi-Lane Spatial Logic for highway scenario reasoning,
with extensions to rural roads~\cite{hilscher2013proving} and urban traffic~\cite{schwammberger2018abstract}.
Alves et al.~\cite{alves2019formalisation} proposed an agent-based architecture to embed the rules of the road into an agent representing the behavior of an autonomous vehicle.
Our proposed Highway Traffic Scenario Logic is inspired by these studies
but is more oriented toward highway trajectory planning.

\subsection{Responsibility Sensitive Safety Model}

The Responsibility Sensitive Safety (RSS) model, introduced by Shalev-Shwartz et al.\cite{shalev2017formal}, offers a set of formal regulations for autonomous vehicles, which have become prevalent in ensuring the safety of autonomous driving.
It has been demonstrated that adherence to RSS rules by all vehicles guarantees the prevention of accidents.
\cite{reimann2024temporal} uses signal temporal logic as a logical formalism to formalize some critical scenarios and use RSS distance for defining danger.
\cite{10421868} presents a formal framework for the formalization of RSS that enables mathematical proof of the safety of control strategies in intersection scenarios.
However, none of these are designed for path planning.
\cite{9575928} incorporates the risk measure into a trajectory planner by utilizing RSS rules and enables autonomous vehicles to plan trajectories with minimal risk.
\cite{9304563} introduces a novel risk-aware RSS approach, which allows for significant reductions in safety margins in a situation-dependent manner to balance between safety and usefulness.
However, directly integrating the RSS into trajectory planning strategies poses a significant challenge.
In practice, many approaches only employ RSS as an additional safety validation step rather than fully incorporating its principles into the trajectory planning process.

\section{Preliminary}
This section provides a formal definition of our Highway Traffic Scenario Logic (HTSL),
which is used for scenario reasoning.

\subsection{Abstract Model}
Our abstract model of HTSL focuses on the multi-lane highway scenarios
containing a uni-directional road with a set of adjacent \textit{lanes} $\mathbb{L}$.
Every car on the road has a unique \textit{car identifier} from the set
$\mathbb{C}$.
At every moment, each car is located in and only in one lane,
and a unique \textit{distance} $d\in \mathbb{D}$ exists between any two cars, with $0 \in \mathbb{D}$.
A partial order relation $\Ldisleq$ is defined on $\mathbb{D}$.
At each step $i$, a unique \textit{action} $a\in \mathbb{A}$ \textit{occurs} to each car,
where the semantic of the step is defined by Linear Temporal Logic (LTL)~\cite{pnueli1977temporal}.
Please note that ``step'' is used here to indicate the order of transitions without physical time connotations.

We can further define a \textit{road} as an arrangement of lanes sorted by their spatial order.
\begin{definition}[Road]
A \textit{road} is an arrangement of $\mathbb{L}$, which is $\mathbf{R}=\langle l^1,l^2,\dots,l^M\rangle$.
A binary relation $\prec_{\text{left}}$ is defined on $\mathbf{R}$ that $l^i \prec_{\text{left}} l^{i+1}$.
\end{definition}

At any given moment, the state of the cars can be abstracted as a \textit{logical snapshot}.
\begin{definition}[Logical Snapshot]
\label{def:logical-snapshot}
A \textit{logical snapshot} $\snap$ is a tuple $(\mathscr{L},\mathscr{D})$, where
\begin{itemize}
    \item[-] $\mathscr{L}: \mathbb{C}\mapsto\mathbb{L}$ is a mapping that assigns a lane to each car.
    \item[-] $\mathscr{D}: \mathbb{C}\times\mathbb{C}\mapsto\mathbb{D}$ is a mapping that assigns a distance to each pair of cars, which will be defined later in Def.~\ref{def:distance}.
\end{itemize}
\end{definition}

A sequence of logical snapshots and actions of each car make up a \textit{logical scenario}.
\begin{definition}[Logical Scenario]
A \textit{logical scenario} $\scen$ is a structure $(\mathbb{C},\mathbb{L},\mathbb{D},\mathbb{A},\langle\snap_0, \mathscr{A}_0,\snap_1,\mathscr{A}_1,\dots\rangle)$, where
\begin{itemize}
    \item[-] $\mathbb{C}$, $\mathbb{L}$, $\mathbb{D}$ and $\mathbb{A}$ have the meanings as above.
    \item[-] $\snap_i=(\mathscr{L}_i,\mathscr{D}_i)$ is the logical snapshot of step $i$.
    \item[-] $\mathscr{A}_i:\mathbb{C}\mapsto\mathbb{A}$ is a mapping that assigns an action of step $i$ to a car.
\end{itemize}
\end{definition}
\begin{definition}[Tail of Logical Scenario]
A \textit{tail} of logical scenario $\scen$, denoted as $\scen_{i:} (i\geq0)$, is the latter part $(\mathbb{C},\mathbb{L},\mathbb{D},\mathbb{A},\langle\snap_i, \mathscr{A}_i,\snap_{i+1},\mathscr{A}_{i+1},\dots\rangle)$ of $\scen$.
\end{definition}

Then, we can provide the syntax and semantics of HTSL.
\begin{definition}[Syntax]
The syntax of the HTSL is defined as follows.
\begin{gather*}
\phi ::= \top \mid \Ldist{c_1}{c_2}{d} \mid \Lon{c}{l} \mid \Lleft{l_1}{l_2} \mid \Loccurs{c}{a} \\
\mid d_1 \Ldisleq d_2 \mid d_1 = d_2 \mid c_1 = c_2 \mid a_1 = a_2 \mid l_1 = l_2 \\
\mid \lnot \phi \mid \phi_1 \to \phi_2 \mid \Next \phi \mid \Globally \phi \mid \forall x\, \phi 
\end{gather*}
where $c,c_1,c_2\in\mathbb{C}$, $l,l_1,l_2\in\mathbb{L}$, $d,d_1,d_2\in\mathbb{D}$, $a,a_1,a_2\in\mathbb{A}$ and $x$ can be any $c$, $l$, $d$ or $a$.
\begin{remark}
We further define $\phi\land \psi$, $\phi\lor \psi$, $\exists x\, \phi$ and $\Finally \phi$ as $(\lnot(\phi \to (\lnot \psi)))$, $((\lnot \phi) \to \psi)$, $(\lnot \forall x (\lnot \phi))$ and $\lnot \Globally \lnot \phi$, respectively, as their definition in first-order logic and modal logic.
\end{remark}
\begin{remark}
The symbols $\Next$, $\Finally$, and $\Globally$ represent ``next'', ``finally'', and ``globally'', respectively, as in the Temporal Logic.
\end{remark}
\end{definition}

\begin{definition}[Semantics]
The satisfaction of formulas w.r.t. a road $\mathbf{R}$ and a logical scenario $\scen$ is defined inductively as follows:
\begin{align*}
    &\mathbf{R},\scen\models \top && & &\text{ for all }\mathbf{R},\scen\\
    &\mathbf{R},\scen\models \Ldist{c_1}{c_2}{d} &\Leftrightarrow & & &\mathscr{D}_0(c_1,c_2)=d\\
    &\mathbf{R},\scen\models \Lon{c}{l} &\Leftrightarrow & & &\mathscr{L}_0(c)=l\\
    &\mathbf{R},\scen\models \Lleft{l_1}{l_2} &\Leftrightarrow & & & l_1 \prec_{\text{left}} l_2\\
    &\mathbf{R},\scen\models \Loccurs{c}{a} &\Leftrightarrow & & & \mathscr{A}_0(c)=a\\
    &\mathbf{R},\scen\models \Next \phi &\Leftrightarrow & & &\mathbf{R},\scen_{1:}\models \phi\\
    &\mathbf{R},\scen\models \Globally \phi &\Leftrightarrow & & & \forall i\geq0,\;\mathbf{R},\scen_{i:}\models \phi
\end{align*}
\end{definition}

\subsection{Action Model}
The action model constrains the transition between logical snapshots.
Considering generality and simplicity, we define the set of actions $\mathbb{A}$ $=$ $\{ \Aincdist,$ $\Adecdist,$ $\Acutl,$ $\Acutr,$ $\Akeep \}$,
which means increase/decrease front distance, left/right cut-int, and passively keep current situation, respectively.
For the formal definition of our action model, please refer to~\ref{action_model}.

\subsection{Safety Constraints}
By defining distance, we integrate the RSS~\cite{shalev2017formal} rules into HTSL as a safety constraint.

\begin{definition}[Distance]
\label{def:distance}
Using $\mathit{loc}(c)$ for the longitudinal coordinate of car $c$ along the road, $\mathit{len}(c)$ for the length of car $c$, and $\mathit{rss}_{c_f}(c)$ for the safe longitudinal distance of car $c$ w.r.t. $c_f$ as its front car. The distance between two cars $c_1$, $c_2$ is $\mathscr{D}(c_1,c_2)\in\mathbb{D}=\{-2,\ldots,2\}$, defined as following:
\begin{equation*}
\mathscr{D}(c_1,c_2) =
\begin{cases}
0, \quad  \rmif \; 0 \leq \mathit{loc}_{12} \leq \mathit{len}_{12},\\
1, \quad \rmif \; \mathit{len}_{12} < \mathit{loc}_{12} \leq \mathit{len}_{12} + \mathit{rss}_{c_1}(c_2),\\
2, \quad \rmif \; \mathit{len}_{12} + \mathit{rss}_{c_1}(c_2) < \mathit{loc}_{12},\\
-\mathscr{D}(c_2,c_1), \quad \rmif \; \mathit{loc}_{12} < 0,
\end{cases}
\end{equation*}
where $\mathit{loc}_{12}=\mathit{loc}(c_1)-\mathit{loc}(c_2)$ and $\mathit{len}_{12}=(\mathit{len}(c_1)+\mathit{len}(c_2))/2$.
The partial order relation $\Ldisleq$ is equivalent to $\leq$ under this definition.
\end{definition}

Then, the RSS rules can be written as follows:
\begin{enumerate}
    \item $\Globally \lnot ( \forall c_1 \forall c_2 \forall d \forall l \, \Ldist{c_1}{c_2}{d} \land 0 \Ldisleq d \Ldisleq 1 \land \Lon{c_1}{l} \land \Lon{c_2}{l} \land \Loccurs{c_2}{\Adecdist})$,
    \item $\Globally \lnot ( \forall c_1 \forall c_2 \forall l_1 \forall l_2 \forall d \, \Ldist{c_1}{c_2}{d} \land -1 \Ldisleq d \Ldisleq 1 \land \Lon{c_1}{l_1} \land \Lon{c_2}{l_2} \land \Lleft{l_1}{l_2} \land \Loccurs{c_2}{\Acutl})$,
    \item $\Globally \lnot ( \forall c_1 \forall c_2 \forall l_1 \forall l_2 \forall d \, \Ldist{c_1}{c_2}{d} \land -1 \Ldisleq d \Ldisleq 1 \land \Lon{c_1}{l_1} \land \Lon{c_2}{l_2} \land \Lleft{l_2}{l_1} \land \Loccurs{c_2}{\Acutr})$,
\end{enumerate}
which means a car should maintain a safe distance from the front car when following and from the front and rear cars when cutting in.

Additionally, a hard constraint of avoiding collision is:
$$
\Globally \lnot ( \forall c_1 \forall c_2 \forall l \forall d \Ldist{c_1}{c_2}{d} \land \Lon{c_1}{l} \land \Lon{c_2}{l} \land d = 0).
$$

\subsection{Goal}
The planning module usually receives a goal from the high-level decision-making layer, which gives instructions like ``lane-change'' or ``overtake''.
\begin{definition}[Goal]
\label{def:goal}
A \textit{goal} $\sigma$ is a formula in the form of
$$\Finally \phi$$
\end{definition}
For example, a goal of ``$c_{ego}$ overtaking all cars'' can be written as 
\begin{equation}
\label{equ:overtake}
    \Finally \forall c \forall d \Ldist{c}{c_{ego}}{d} \land \lnot (c = c_{ego}) \land d \Ldisleq 0.
\end{equation}

\section{Method}
\begin{figure}[t]
\centerline{\includegraphics[width=0.48\textwidth]{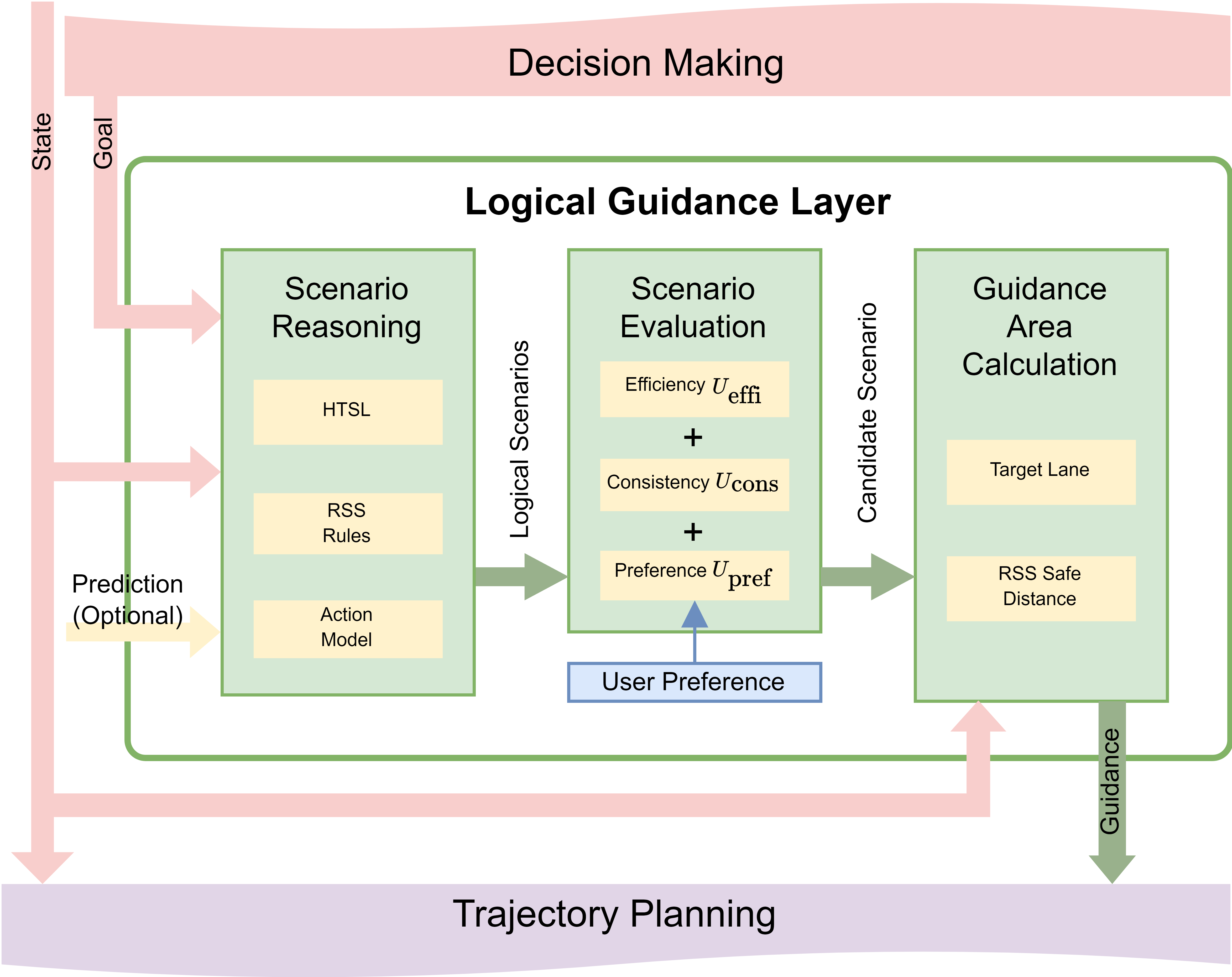}}
\caption{Our LGL (green) is inserted between decision-making (red) and trajectory-planning (purple).
The input includes the goal obtained from decision-making and the current state.
An extra prediction input of environmental vehicles' actions is also acceptable.
The LGL finally generates the guidance area in the S-L-T-V space with the consideration of efficiency, consistency, and user preferences.
}
\label{fig:mainfig}
\end{figure}

The proposed Logical Guidance Layer (LGL), delineated in Fig.~\ref{fig:mainfig}, comprises three key components: the scenario reasoning module, the scenario evaluation module, and the guidance area calculation module.
In essence, the LGL serves to identify local target regions for the trajectory planning layer,
aiming to optimize safety and traffic efficiency and meet user preferences.

The scenario reasoning module operationalizes a goal extracted from the decision-making layer into the structured format defined by Definition \ref{def:goal}.
This involves translating the current state into a road structure $\mathbf{R}$ and a logical snapshot $\snap_0$. 
Then, the scenario reasoning module anticipates potential future logical scenarios to satisfy RSS safety constraints, 
and the scenario evaluation module prioritizes these scenarios based on a utility function that accounts for efficiency, consistency, and user preferences. 
Subsequently, the most advantageous logical scenario, determined through this evaluation process, is delineated within a guidance area situated in the S-L-T-V space.
This guidance area is defined by constraints encompassing longitudinal coordinates ($s$), lateral coordinates ($l$), time ($t$), and target velocity ($v$).

\subsection{Scenario Reasoning}
Formally, the scenario reasoning module receives a goal $\sigma$ from the decision-making layer and generates a set of possible logical scenarios $\mathbb{S}=\{\scen_1,\scen_2,\dots\}$ that satisfy the goal and all constraints with the HTSL.

\paragraph{Initial Condition}
The HTSL takes the road structure $\mathbf{R}$ and current logical snapshot $\snap_0$ as the initial condition,
which can be easily obtained from the observation.
Technically, to reduce the complexity of solving, the set of cars $\mathbb{C}$ is narrowed to the ego vehicle and its surrounding cars on every lane.

\paragraph{Action}

By default, the ego vehicle can take all actions defined in Def.~\ref{def:action}.
The environmental vehicles are assumed to only take the $\Akeep$ action, except
when an extra prediction is provided.
The action of environmental vehicles can be identified by an external perception module (for example, by their turn signal lights), which is beyond the scope of this paper.

\paragraph{Solver}
We use \textit{Telingo}~\cite{cabalar2019telingo} to implement the HTSL and solve the scenario reasoning problem.
Due to the variety of traffic scenarios, the number of logical scenarios is enormous.
We address this problem by projecting the answer set onto the next logical snapshot $\snap_1$, which only has tens of different results.
Since the logical snapshot $\snap_1$ is used for calculating the guidance area in Section~\ref{subsec:guide-generate},
different scenarios with the same logical snapshot $\snap_1$ have no difference in the generated guidance area.

\subsection{Scenario Evaluation}

The logical scenarios introduced by the scenario reasoning module are ranked based on their utility function to seek higher efficiency and meet user preferences.
\begin{equation}
    U(\scen) = w_1 U_{\text{effi}}(\scen) + w_2 U_{\text{perf}}(\scen) + w_3 U_{\text{cons}}(\scen)
\end{equation}
where $w_1$, $w_2$ and $w_3$ are weights adjusted through experimentation.

\paragraph{Efficiency Evaluation}

Since all candidate logical scenarios end with the goal being satisfied,
the length of the scenario can be used as a metric for efficiency, i.e.

\begin{equation}
    U_{\text{effi}}(\scen) = - \mathrm{len}(\scen)
\end{equation}

\paragraph{User-preference Evaluation}
User preferences can be diverse, 
so we use logical formulae to describe user preferences,
which ensures maximum freedom of personalization.
Formally, a user preference is a group of logical formulae in the form of
\begin{equation*}
    \phi \to \mathsf{pref}(u)
\end{equation*}
where $\phi$ is a HTSL formula and $u$ is an integer.
The predicate $\mathsf{pref}/1$ extends the HTSL,
to indicate the degree of user preference for the logical scenario.
Then, the user preference utility is defined by the sum of $u$ that is derived from $\scen$:
\begin{equation}
    U_{\text{perf}}(\scen) = \sum u \;\text{for each}\;\mathsf{perf}(u)
\end{equation}
For example, a user preference stating ``hate of overtaking from the right'' could be expressed as
\begin{multline*}
    \Finally (\forall c \forall d \forall l_1 \forall l_2 \, \Ldist{c}{c_{ego}}{d} \land 1 \Ldisleq d \land \Lon{c}{l_1} \land \\ 
    \Lon{c_{ego}}{l_1} \land \Finally (\Ldist{c}{c_{ego}}{0}) \land \\
    \Lon{c}{l_1} \land \Lon{c_{ego}}{l_2} \land \Lleft{l_1}{l_2} ) \to \mathsf{perf}(-1)
\end{multline*}

\paragraph{Consistency Evaluation}

To maintain consistency with the scenario selected in the preceding step, 
the size of the intersection between the guidance area of the current step and that of the previous step is calculated as a measure of consistency.
\begin{equation}
    U_{\text{cons}}(\scen) = \mu(g(\scen) \cap g(\scen'))
\end{equation}
where $\scen'$ is the logical scenario selected in the previous step,
$g(\cdot)$ is the guidance area 
and $\mu(\cdot)$ represents the measure of a set in the S-L-T-V space.
The calculation of the guidance area is introduced in Section~\ref{subsec:guide-generate}.

\subsection{Guidance Area Calculation}
\label{subsec:guide-generate}
We use the next logical snapshot $\snap_1$ of a logical scenario $\scen$ for calculating the guidance area.
According to Definition~\ref{def:logical-snapshot},
a logical snapshot includes the lanes of vehicles and the distances between them.
This can be mapped into the S-L-T-V space as a guidance for the following trajectory planning, 
where S and L are coordinates under the Frenet frame, and T and V are time and target speed.

\paragraph{L-axis}
The L-axis describes the lane of a vehicle,
thus we can take the lane of the ego vehicle $\mathscr{L}_1(c_{ego})$
and map into a region of target l-coordinates straightforwardly.

\paragraph{S-axis, T-axis, and V-axis}

The distances between the ego vehicle and the others describe the constraints of the S-coordinate, time, and target speed.
The RSS safe longitudinal distance is~\cite{shalev2017formal}:
\begin{equation}
\label{equ:rss}
rss_{c_f}(c_r) = \max\left[v_r\rho + \frac{1}{2}a_1\rho^2+\frac{(v_r+\rho a_1)^2}{2a_2} - \frac{v_f^2}{2a_3}, 0\right]
\end{equation}
where $v_f$ is the longitudinal speed of the front car $c_f$, $v_r$ is the longitudinal speed of the rear car $c_r$, and $\rho$, $a_1$, $a_2$ and $a_3$ are constants.
Combined with Definition~\ref{def:distance}, we can obtain the range of s-coordinates for different times and target speeds.
The region mapped by the distances between the ego vehicle and different vehicles takes the intersection, which is the final guidance area.
An empty intersection indicates an invalid guidance area that will be discarded.

\section{EXPERIMENT AND RESULTS}

\begin{figure*}[t]
\centering
	\begin{subfigure}[b]{0.98\textwidth}
		\includegraphics[width=\linewidth]{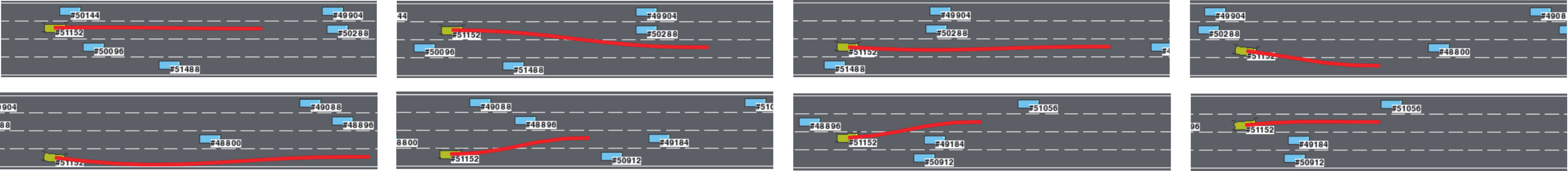}
		\caption{LGL+Lattice}
		\label{fig:sub1}
	\end{subfigure}
	\begin{subfigure}[b]{0.995\textwidth}
		\includegraphics[width=\linewidth]{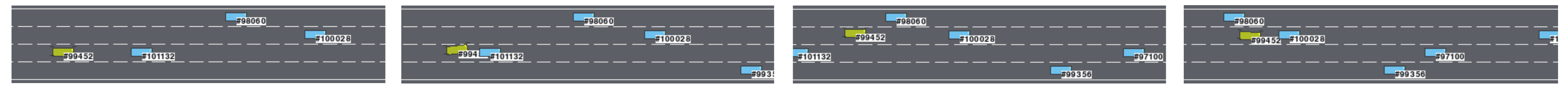}
		\caption{Lattice}
		\label{fig:sub2}
	\end{subfigure}
	\begin{subfigure}[b]{0.995\textwidth}
		\includegraphics[width=\linewidth]{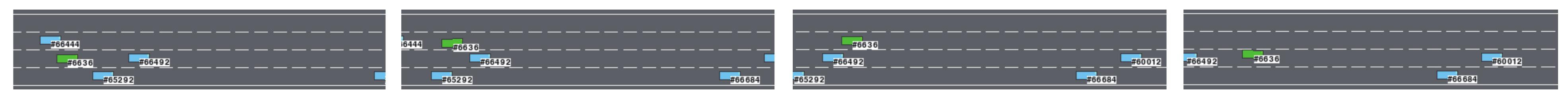}
		\caption{RL}
		\label{fig:sub3}
	\end{subfigure}
	\begin{subfigure}[b]{0.995\textwidth}
		\includegraphics[width=\linewidth]{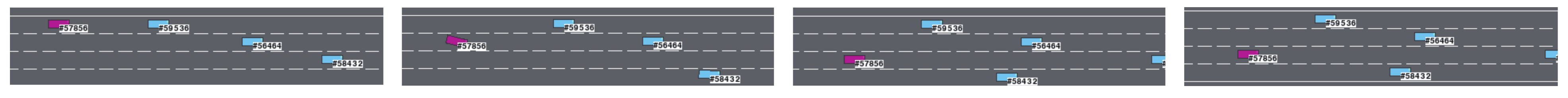}
		\caption{IDM+MOBIL}
		\label{fig:sub4}
	\end{subfigure}
\caption{Demonstration of the operation of vehicles controlled by four algorithms in high-density environments. (a) Results of the LGL+Lattice (left to right, top to bottom). The red lines represent the planned trajectories. (b) Results of the pure Lattice algorithm. (c) Results of the RL algorithm. (d) Results of IDM+MOBIL.}
\label{fig:screenshot}
\end{figure*}

\begin{table}[tbp]
\centering
\resizebox{\columnwidth}{!}{
\begin{tabular}{llcc|cc|c}
\toprule
\multirow{2}{*}{\textbf{Dens.}} & \multirow{2}{*}{\textbf{Method}} & \multicolumn{2}{c|}{\textbf{Efficiency}} & \multicolumn{3}{c}{\textbf{Safety}}        \\ \cmidrule{3-7}
                  &                   & \multicolumn{2}{c|}{\textbf{Ave. Speed}} & \multicolumn{2}{c|}{\textbf{Min. TTC}}    & \textbf{Succ.\%}  \\
\midrule
\multirow{4}{*}{High} & LGL+Lattice & 24.75 & (1.80) & 1.98 & (1.20) & 100.0  \\
                      & Lattice     & 24.82 & (3.49) & 0.52 & (0.73) & 87.5  \\
                      & RL          & 27.86 & (2.03) & 0.83 & (0.59) & 73.4  \\
                      & IDM+MOBIL   & 21.94 & (0.94) & 11.76 & (5.10) & 100.0  \\
\midrule
\multirow{4}{*}{Low} & LGL+Lattice  & 28.42 & (1.23) & 3.06 & (2.61) & 100.0  \\
                      & Lattice     & 28.37 & (2.56) & 2.86 & (3.41) & 90.6  \\
                      & RL          & 29.14 & (1.12) & 1.61 & (0.92) & 100.0  \\
                      & IDM+MOBIL   & 25.80 & (1.42) & 16.15 & (3.57) & 100.0  \\
\bottomrule
\end{tabular}
}
{\raggedright Note: Data in the table are presented as mean (standard deviation). \par}
\caption{Comparison of Our Method with Baselines on Efficiency and Safety}
\label{tab::1}
\end{table}

\begin{figure}[htbp]
    \centering
    \includegraphics[width=0.9\linewidth]{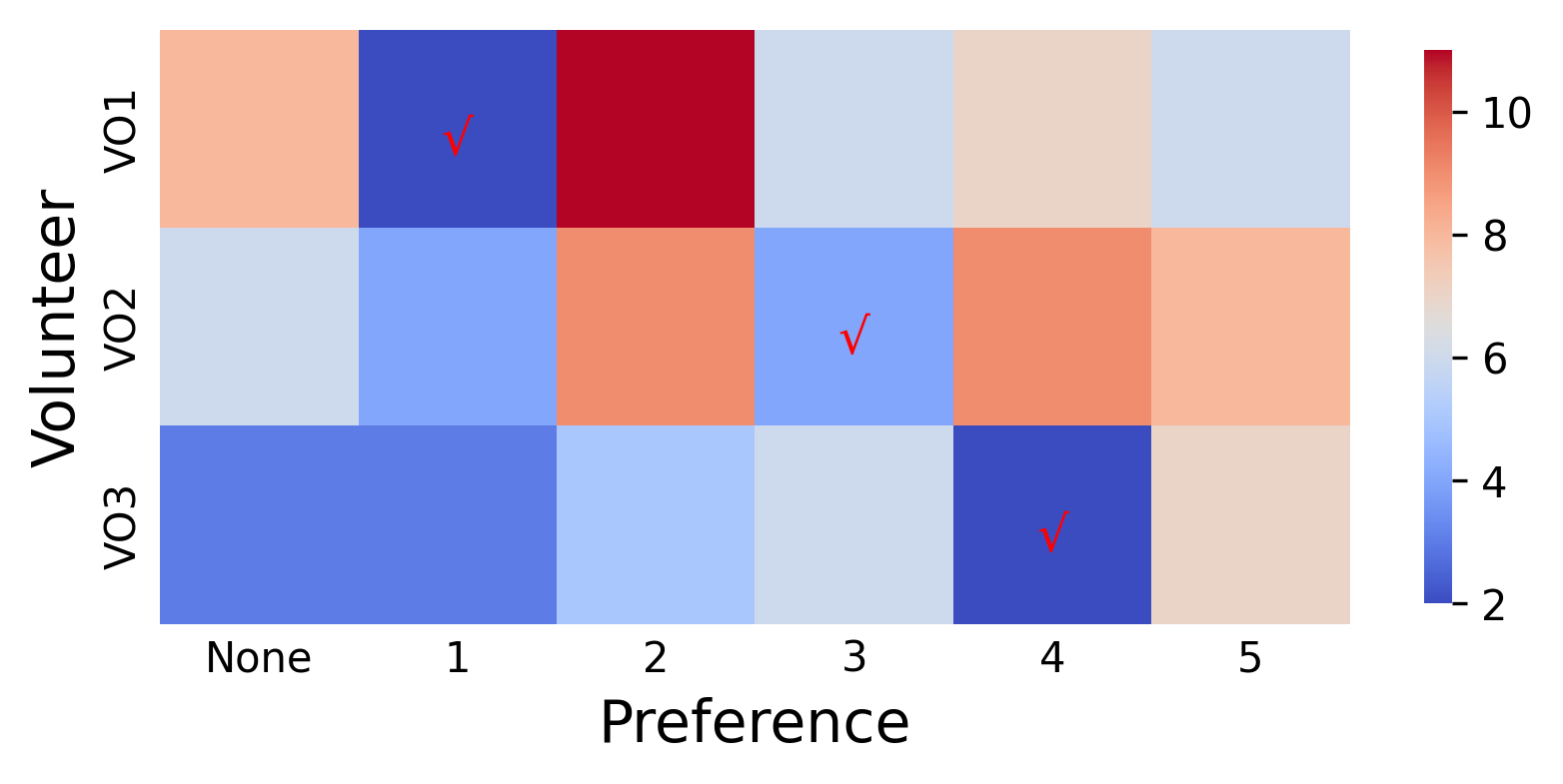}
    \caption{The number of overrides of 3 volunteers under each user preference. The volunteers' choices are marked by red check marks.}
    \label{fig:pref}
\end{figure}

This section experimentally demonstrates the efficacy of our proposed LGL in terms of efficiency, safety, and alignment with user preferences.

\subsection{Efficiency and Safety}

\paragraph{Simulation Setting}
We use HighwayEnv~\cite{highway-env} as a simulator for simulating a four-lane highway scenario.
We vary traffic density across two settings: high (vehicle density is 1.0) and low (vehicle density is 0.5).
To enhance the fidelity of our simulations, we employ the Naturalistic Driving Environment (NDE) as introduced in~\cite{yan2021distributionally} to emulate the natural driving behaviors observed in human-operated vehicles.
The goal for LGL is settled to Equ.~(\ref{equ:overtake}).
Trajectory generation at a local level is facilitated by a Lattice planner,
while the determination of acceleration and steering angle is accomplished through the utilization of the pure pursuit algorithm~\cite{coulter1992implementation}.
Notably, user preferences and predictions are excluded from the experimental setup for efficiency and safety.

\paragraph{Methods for Comparison}
To demonstrate the role of our proposed LGL in improving the efficiency and safety,
the LGL is compared with:
\begin{itemize}
    \item \textbf{Lattice}: a naive Lattice planner with all hyper-parameters the same as the one used by LGL, but with a default sample area;
    \item \textbf{RL}: a DQN-based Reinforcement Learning agent presented by \cite{stable-baselines3}, trained for 200k steps;
    \item \textbf{IDM+MOBIL}: a simple vehicle controlled by IDM\cite{treiber2000idm} and MOBIL\cite{kesting2007mobil}.
\end{itemize}

\paragraph{Metrics}
All algorithms to be
compared are evaluated using the following metrics:
\begin{itemize}
    \item \textbf{Efficiency}: Efficiency is measured by the average speed of the vehicle, with a target speed of $30\mathrm{m/s}$;
    \item \textbf{Safety}: We computed the ego vehicle's minimum Time-To-Collision (TTC) during each 60-second episode and the success rate of completing the whole episode.
\end{itemize}
\subsection{User Preference}

We provide the following five user preferences:
\begin{enumerate}[itemindent=3em]
    \item[\textbf{Pf 1:}] hate overtaking from the right;
    \item[\textbf{Pf 2:}] hate overtaking from the left;
    \item[\textbf{Pf 3:}] prefer the rightmost lane;
    \item[\textbf{Pf 4:}] prefer the leftmost lane;
    \item[\textbf{Pf 5:}] hate being sandwiched between two cars.
\end{enumerate}

Three volunteers were asked to choose one of the user preferences among them
and watch ten episodes of each preference for one minute each,
without knowing which preferences were being used.
The volunteers could override the ego vehicle at any time using the keyboard.
The volunteers' preference choices and the number of overrides under each preference are shown in Fig.~\ref{fig:pref}.

\subsection{Results and Analysis}

The comparison results for efficiency and safety are summarized in Table \ref{tab::1}, where we conducted 64 episodes of testing for each algorithm, respectively, in different traffic densities.
Our proposed LGL achieves a good balance between efficiency and safety, showing a clear advantage over other algorithms.
Compared with the lattice algorithm without LGL, 
LGL significantly improves safety while maintaining high efficiency,
and the TTC is more in line with the ``two-second rule'', a rule of thumb well-known to human drivers.
The RL algorithm always behaves more aggressively and is substantially less safe in high-density environments.
IDM+MOBIL algorithm, on the other hand, is overly conservative, sacrificing efficiency for safety.
Fig.~\ref{fig:screenshot} demonstrates the behavior of all methods in the high-density environment, showing that the safe distance between the front and rear is well controlled under the guidance of the LGL.

The experiment results for user preference are shown in Fig.~\ref{fig:pref}.
The results show that LGL is significantly effective in satisfying different user preferences and reducing the frequency of user takeovers.
This suggests the potential of the LGL to improve the experience of different users.

\section{Conclusion}

This paper introduces a novel component for autonomous driving trajectory planning in highway environments, termed the Logical Guidance Layer (LGL).
It accepts goals from the decision-making layer, infers potential future scenarios through the scenario reasoning module, and selects an optimal candidate via the scenario evaluation module.
Subsequently, it generates guidance regions in the S-L-T-V space to accommodate major local trajectory planning algorithms.
The Responsibility Sensitive Safety model ensures safety throughout the process,
and user preferences defined by logical formulae can be considered during the scenario evaluation.
Experimental results demonstrate that LGL significantly enhances safety while maintaining high efficiency
and offers customizable user preferences to enhance the user experience of autonomous driving.
Future work includes extending LGL to urban traffic scenarios and exploring its applications in improving human-machine interaction experiences in autonomous driving.

\addtolength{\textheight}{-0cm}   



\section*{APPENDIX}
\newcounter{appendix}
\renewcommand{\theappendix}{Appendix \Alph{appendix}}
\subsection{Action Model of HTSL}
\refstepcounter{appendix}
\label{action_model}
\begin{definition}[Action Model]
\label{def:action}
The set of action $\mathbb{A}$ $=$ $\{ \Aincdist,$ $\Adecdist,$ $\Acutl,$ $\Acutr,$ $\Akeep \}$
is defined by following rules.
\begin{enumerate}
    \item $\forall c_1 \forall c_2 \forall d_1 \exists d_2 \Loccurs{c_1}{\Aincdist} \land \Ldist{c_2}{c_1}{d_1} \land 0 \Ldisleq d_1 \land d_1 \Ldisleq d_2 \to \Next \Ldist{c_2}{c_1}{d_2}$ \\ \textit{($\Aincdist$ may increase the distance to the front)}
    \item $\forall c_1 \forall c_2 \forall d_1 \exists d_2 \Loccurs{c_2}{\Aincdist} \land \Ldist{c_2}{c_1}{d_1} \land 0 \Ldisleq d_1 \land d_2 \Ldisleq d_1 \to \Next \Ldist{c_2}{c_1}{d_2}$ \\ \textit{($\Aincdist$ may decrease the distance to the rear)}
    \item $\forall c \forall l \Loccurs{c}{\Aincdist} \land \Lon{c}{l} \to \Next \Lon{c}{l}$ \\ \textit{($\Aincdist$ keeps current lane)}
    \item $\forall c_1 \forall c_2 \forall d_1 \exists d_2 \Loccurs{c_1}{\Adecdist} \land \Ldist{c_2}{c_1}{d_1} \land 0 \Ldisleq d_1 \land d_2 \Ldisleq d_1 \to \Next \Ldist{c_2}{c_1}{d_2}$
    \item $\forall c_1 \forall c_2 \forall d_1 \exists d_2 \Loccurs{c_2}{\Adecdist} \land \Ldist{c_2}{c_1}{d_1} \land 0 \Ldisleq d_1 \land d_1 \Ldisleq d_2 \to \Next \Ldist{c_2}{c_1}{d_2}$
    \item $\forall c \forall l \Loccurs{c}{\Adecdist} \land \Lon{c}{l} \to \Next \Lon{c}{l}$
    \item $\forall c \forall l_1 \forall l_2 \Loccurs{c}{\Acutl} \land \Lon{c}{l_1} \land \Lleft{l_2}{l_1} \to \Next \Lon{c}{l_2}$
    \item $\forall c_1 \forall c_2 \forall d \Loccurs{c_1}{\Akeep} \land \Ldist{c_2}{c_1}{d} \land \lnot \Loccurs{c_2}{\Aincdist} \land \lnot \Loccurs{c_2}{\Adecdist} \to \Next \Ldist{c_2}{c_1}{d}$ \\ \textit{($\Acutl$ change distance passively)}
    \item $\forall c \forall l_1 \forall l_2 \Loccurs{c}{\Acutr} \land \Lon{c}{l_1} \land \Lleft{l_1}{l_2} \to \Next \Lon{c}{l_2}$ \\ \textit{($\Acutr$ will switch to the right lane)}
    \item $\forall c_1 \forall c_2 \forall d \Loccurs{c_1}{\Acutr} \land \Ldist{c_2}{c_1}{d} \land \lnot \Loccurs{c_2}{\Aincdist} \land \lnot \Loccurs{c_2}{\Adecdist} \to \Next \Ldist{c_2}{c_1}{d}$
    \item $\forall c_1 \forall c_2 \forall d \Loccurs{c_1}{\Akeep} \land \Ldist{c_2}{c_1}{d} \land \lnot \Loccurs{c_2}{\Aincdist} \land \lnot \Loccurs{c_2}{\Adecdist} \to \Next \Ldist{c_2}{c_1}{d}$
    \item $\forall c \forall l \Loccurs{c}{\Akeep} \land \Lon{c}{l} \to \Next \Lon{c}{l}$
\end{enumerate}
\begin{remark}
All rules above omit $\lnot(c_1 = c_2)$ for brevity.
\end{remark}
\end{definition}



\bibliographystyle{IEEEtran}
\bibliography{IEEEabrv,reference}

\end{document}